\documentclass{article}
\usepackage{ijcai26}

\usepackage{times}
\usepackage{soul}
\usepackage{url}
\usepackage[hidelinks]{hyperref}
\usepackage[utf8]{inputenc}
\usepackage[small]{caption}
\usepackage{graphicx}
\usepackage{amsmath}
\usepackage{amsthm}
\usepackage{booktabs}
\usepackage{adjustbox}
\usepackage{array}
\usepackage{algorithm}
\usepackage{algorithmic}
\usepackage[switch]{lineno}
\usepackage{xcolor}

\title{EmbeWebAgent: Embedding Web Agents into Custom Enterprise UIs}

\author{
Chenyang Ma$^{1,2}$\thanks{Work done during an internship at IBM.},
Clyde Fare$^{1}$,
Matthew Wilson$^{1}$,
Dave Braines$^{1}$
\affiliations
$^1$IBM Research Europe, UK \quad $^2$University of Oxford
\emails
dannymcy@umich.edu
\\[1.0em]
\texttt{\href{https://youtu.be/Cy06Ljee1JQ}{\textcolor{magenta}{Demo Video}}}
}

\newcommand{\sys}{EmbeWebAgent}

\begin{document}

\maketitle

\begin{abstract}
Most web agents operate at the human interface level, observing screenshots or raw DOM trees without application-level access, which limits robustness and action expressiveness. In enterprise settings, however, explicit control of both the frontend and backend is available. We present \textbf{\sys}, a framework for embedding agents directly into existing UIs using lightweight frontend hooks (curated ARIA and URL-based observations, and a per-page function registry exposed via a WebSocket) and a reusable backend workflow that performs reasoning and takes actions. \sys\ is stack-agnostic (e.g., React or Angular), supports mixed-granularity actions ranging from GUI primitives to higher-level composites, and orchestrates navigation, manipulation, and domain-specific analytics via MCP tools. Our demo shows minimal retrofitting effort and multi-step behaviors grounded in a live UI setting. 
Demo video is available at {\href{https://youtu.be/Cy06Ljee1JQ}{\textcolor{magenta}{this link}}}.
\end{abstract}

\section{Motivation and Contributions}

\begin{figure}[t]
  \centering
  \includegraphics[width=\linewidth]{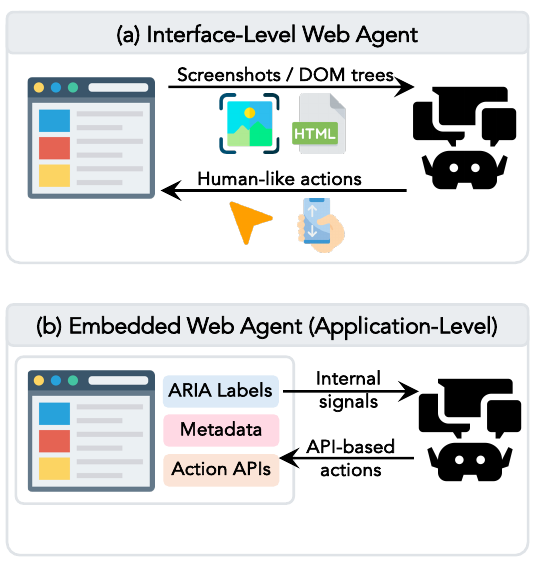}
  \vspace{-4mm}
  \caption{\textbf{(a) Interface-level agent} (screenshots/DOM trees, simulated clicks) vs.\ \textbf{(b) Embedded agent (\sys)} with ARIA labels as observations and explicit UI actions.}
  \label{fig:teaser}
  \vspace{-0.7em}
\end{figure}

Recent advances in large language models (LLMs) and large multimodal models (LMMs)~\cite{GPT-4o_openai,llama3_Dubey} have catalyzed research on autonomous web agents across coding, system operation, and research tooling~\cite{ning2025survey}. Most web agents operate at the interface level, perceiving content via screenshots~\cite{ZhengGK0024,GouWZXCS0025,ma2024spatialpin} or Document Object Model (DOM) trees~\cite{DengGZCSWSS23,0036YZXLL0DMYZ024}, and simulating human-like actions (e.g., clicks, text entry) on visible UI elements without access to underlying application logic, as shown in Fig.~\ref{fig:teaser}. These agents typically rely on proprietary, state-of-the-art LLMs/LMMs as their backbone and often struggle to reliably execute complex workflows. In enterprise settings, however, organizations have explicit control of both the frontend and backend, and richer internal signals such as Accessible Rich Internet Applications (ARIA) labels, action APIs, and page metadata can be made available by embedding agents directly into applications. Using such signals can reduce the complexity of the tasks the agent needs to accomplish, hence improving performance and robustness. 

Yet, few frameworks explicitly target embedding agents into existing web applications~\cite{copilotkit,superinterface}, and directly adapting them is often impractical in enterprise settings, where legacy UIs are already deployed and built on heterogeneous frontend web stacks (e.g., React, Angular). Specifically, these limitations include tight coupling to a specific web stack, heavy per-page instrumentation that does not generalize (e.g., hardcoded actions and observations), and entanglement of agent logic with UI code, making retrofitting costly and backend reuse difficult. Emerging proposals such as WebMCP~\cite{webmcp2026} define only how pages expose tools; how an agent observes the UI and orchestrates them is left open.

\begin{figure*}[t]
  \centering
  \includegraphics[width=\linewidth]{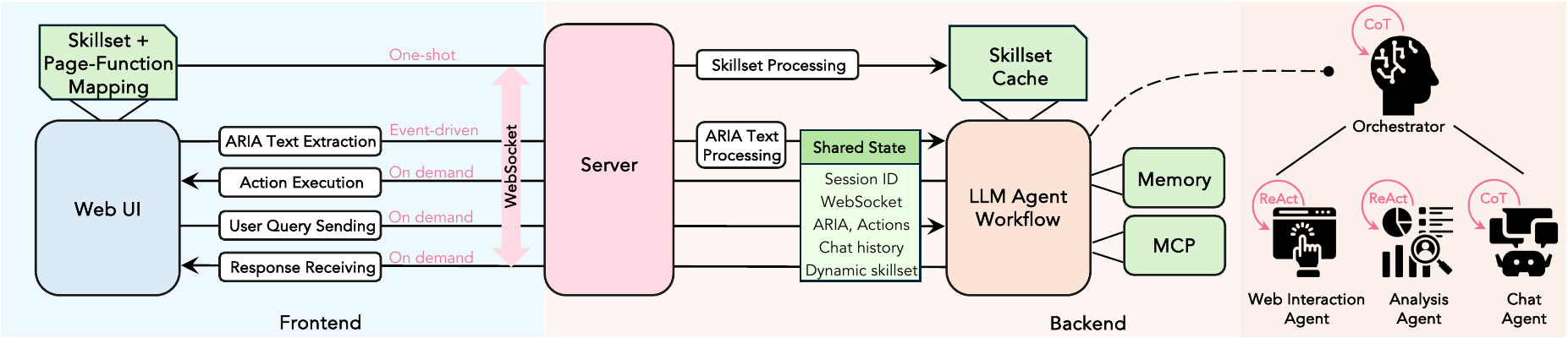}
  \vspace{-4mm}
  \caption{\textbf{\sys\ pipeline.} The frontend shim exposes curated ARIA observations and a per-page function registry via a WebSocket. The backend maintains session state, filters actions according to the current page, and coordinates a multi-agent workflow to infer navigation and manipulation actions, as well as invoke domain tools (via MCP or web APIs).}
  \label{fig:pipeline}
  \vspace{-0.7em}
\end{figure*}

Thus we present \textbf{\sys} defined by the following design principles:
\begin{itemize}
\item \textbf{An Embedded Agent Framework.} Curated ARIA labels and URL-based observations, explicit exposure of navigation URLs, and a per-page function registry that restricts actions to those valid for the current page.
\item \textbf{Minimal, Stack-Agnostic Integration.} A small frontend shim, while keeping all agent reasoning and workflows in the backend.
\item \textbf{Mixed-Granularity Actions.} A combination of GUI primitives (click, type, select, scroll) and higher-level composite actions (e.g., add item to list), avoiding per-page prompt engineering; this is related in spirit to hybrid GUI/API agents~\cite{SongXZN25,api_gui_zhang,lu2025kitchenvla,ma2025coopera,tu2026unitac,kim2026framing}.
\item \textbf{Multi-Agent Orchestration.} The web-interaction agent acts alongside an analysis agent that invokes domain tools via the Model Context Protocol (MCP) in our demo, followed by a chat summarizer that uses lightweight Chain-of-Thought (CoT)~\cite{Wei0SBIXCLZ22}. User-specific interactions are achieved via a session-scoped shared state.
\end{itemize}

Our demo showcases \textbf{\sys} in a chemistry-based application, where the agent navigates the interface, manipulates views, and invokes analytics, illustrating end-to-end behavior with minimal frontend retrofitting.

\section{System Overview}
In \textbf{\sys}, we assume full access to the frontend codebase. The agent handles tasks of varying complexity, including nested navigation across URLs, interface manipulation (e.g., add/remove entities, edit visuals), and chatting with retrieved knowledge/tools. The overall architecture and design principles are shown in Fig.~\ref{fig:pipeline}. The frontend and backend have clearly separated responsibilities, described below.

\paragraph{Frontend Shim (Minimal Retrofitting).} The frontend contains lightweight hooks that expose structured observations and function tools to the backend. Specifically, it provides: (i) \emph{Observations}: curated ARIA labels (as used in prior web agent work~\cite{glek2023gptaria,WILBUR_Lutz,agent-e_Abuelsaad}) and the current page URL; (ii) \emph{Action surface}: a per-page function registry specifying function names, schemas, and page associations; and (iii) \emph{Communication channel}: a WebSocket interface for bidirectional events (queries, observations, actions, and chat messages). In our demo, the frontend shim requires $\sim$150 lines of code, and the function registry requires $\sim$200 lines.

\paragraph{Backend (Reasoning \& Control).} The backend handles reasoning, planning, and action inference. A session-scoped state maintains the latest observations, the page-filtered function set corresponding to the current URL, recent chat history, and previously executed actions. An orchestrator routes user goals to three coordinated agents: (i) a \emph{web-interaction agent}, which infers navigation and manipulation actions from the filtered function registry; (ii) an \emph{analysis agent}, which invokes domain-specific tools via MCP; and (iii) a \emph{chat agent}, which provides concise user feedback. Web and analysis agents follow a ReAct-style loop~\cite{YaoZYDSN023}; the chat agent uses CoT~\cite{Wei0SBIXCLZ22}. Actions are returned to the frontend via the shim, and resulting UI updates are streamed back as new observations, thereby closing the interaction loop.

\section{Architecture Details}

\paragraph{Observations via ARIA Labels + Page URL.}
Rather than relying on screenshots or full DOM trees, we observe the interface
through ARIA labels paired with the
current page URL. ARIA is a W3C standard for labeling interactive elements,
providing semantic, human-readable cues that are well suited for LLM reasoning~\cite{glek2023gptaria,WILBUR_Lutz,agent-e_Abuelsaad}. As we
have access to the frontend codebase, ARIA labels can be selectively attached
to meaningful components (e.g., a button's purpose, a link's destination) and
paired with their HTML tags for additional context. On the backend, observations are filtered by tag type to discard decorative or irrelevant elements (e.g., icons, layout wrappers) before being passed to agents.

\paragraph{Mixed-Granularity Actions and Explicit Navigation.}
Actions are defined as frontend functions annotated with purpose descriptions
and parameter schemas, forming a reusable skillset. These annotations serve a
dual role: documenting behavior for developers and providing affordances to guide an LLM-based tool-calling capability. Only metadata (names, schemas, page associations) is sent to
the backend for caching; implementations and execution remain on the frontend,
preserving clean separation. A key design choice is mixed granularity: low-level GUI primitives (e.g., click, scroll, type) support simple and general manipulations, while higher-level composites (e.g., add-item-to-list)
encapsulate multi-step workflows, avoiding per-page prompt engineering.
Navigation is treated as an explicit action: ARIA-labeled links are extracted
and incorporated into a dedicated navigation function that exposes all
reachable URLs to the agent, enabling robust handling of nested and multi-level navigation. To constrain the action space, the function registry is paired with a page-function mapping so that only functions relevant to the current URL are visible to the agent, reducing spurious calls. The design of mixed-granularity actions is similar in spirit to hybrid GUI/API agents~\cite{SongXZN25,api_gui_zhang,lu2025kitchenvla,ma2025coopera,tu2026unitac,kim2026framing} but tailored to embedded enterprise settings where backend caching and reusable abstractions are critical for scalability.

\paragraph{Lightweight Communication.}
Frontend-backend communication uses a WebSocket with event-driven triggers to
minimize overhead. Chat messages and action requests are exchanged on demand,
while the skillset and page-function mapping are sent once at page load.
Observations are pushed lazily: only when the URL changes or UI elements are
modified by the agent or user.

\paragraph{Session Grounding.}
An in-memory shared state is maintained per user session, serving three
purposes: (i) isolating concurrent users by scoping observations, actions, and
chat history to a session ID; (ii) storing the latest DOM observations and
agent-generated actions, updated via a WebSocket upon action execution, ensuring reasoning remains grounded in the state of the live interface; and (iii) maintaining a dynamic skillset filtered from the page-function mapping based on the current URL. A lightweight memory of the last $n$ chat turns maintains context across multi-step interactions.

\section{Demo Scenario}

\begin{figure}[t]
  \centering
  \includegraphics[width=\linewidth]{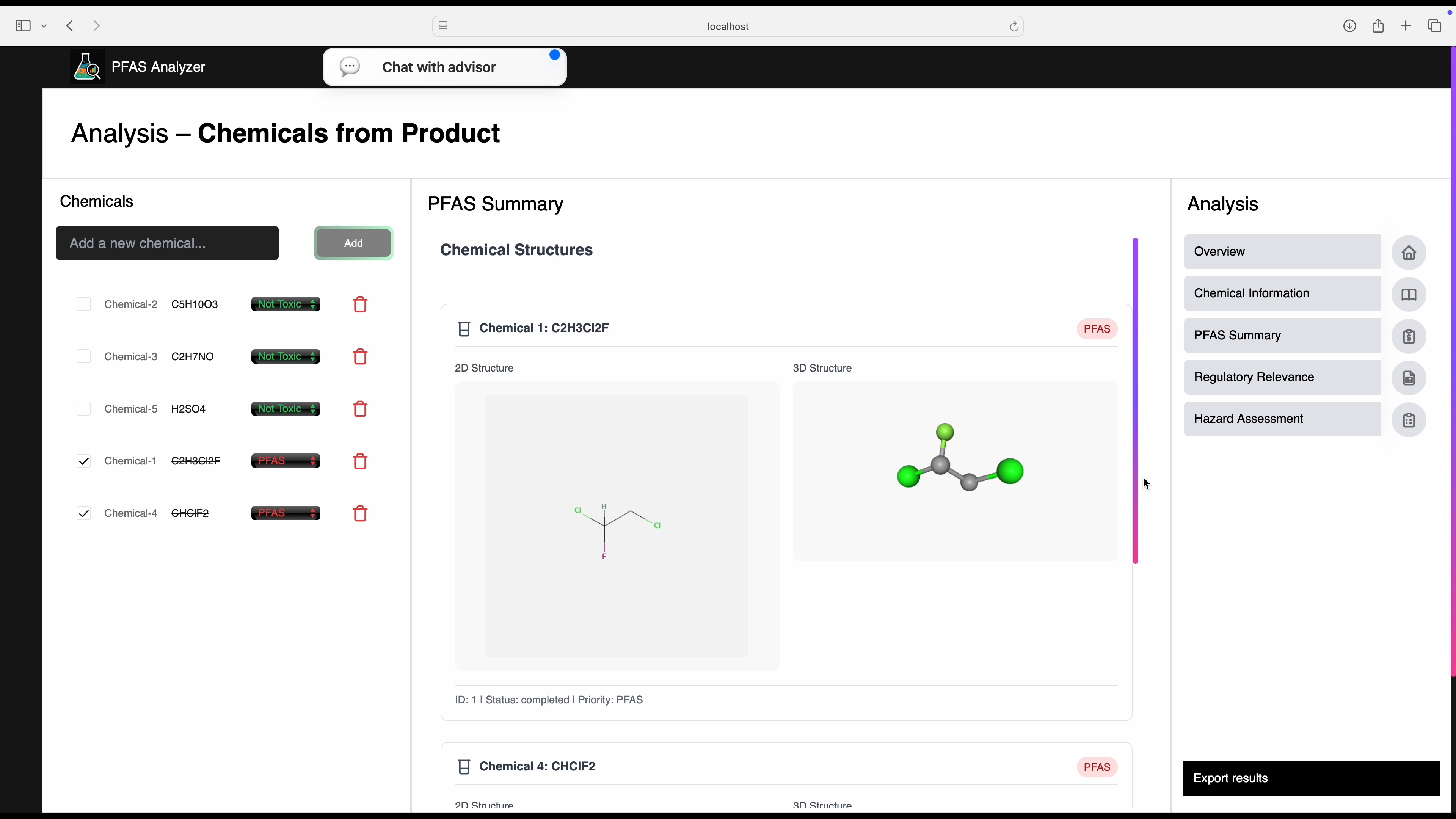}
  \vspace{-4mm}
  \caption{\textbf{One page of the chemistry analysis UI} used in our demo.}
  \label{fig:demo}
  \vspace{-0.7em}
\end{figure}

The demo features a chemistry analysis UI where users investigate whether a
compound is a Per- and Polyfluoroalkyl Substance (PFAS)~\cite{Buck2011-og,wang2025calcium},
synthetic chemicals widely used in consumer products that persist in the
environment and accumulate in the human body. One page of the UI is shown in Fig.~\ref{fig:demo}. A walkthrough proceeds as follows:    
\begin{enumerate}
\item \textbf{Goal input}: The user asks ``Check if this SMILES is a PFAS and
generate a short report.'' SMILES (Simplified Molecular-Input
Line-Entry System) is a line notation that encodes molecular structure
as a text string.
\item \textbf{Grounding}: The frontend sends the ARIA snapshot (search box,
``Analyze'' button, ``Reports'' tab) and the current page's function registry
(e.g., \texttt{type(textField, value)}, \texttt{click(analyze)},
\texttt{navigate(url)}).
\item \textbf{Action inference}: The web-interaction agent types the SMILES string, triggers analysis, and navigates to ``Reports.''
\item \textbf{Analytics}: The analysis agent calls a PFAS classifier via MCP and retrieves supporting evidence (the architecture generalizes to any tool exposed via MCP or web APIs).
\item \textbf{Response}: The chat agent summarizes results; the user sees both
the updated UI and a concise explanation.
\end{enumerate}

\section{Implementation}
We have demonstrated that the architecture works with both React and Angular
frontend frameworks, validating stack-agnostic integration; the live demo
uses React. The agent
workflow is implemented with DSPy~\cite{khattab2024dspy} using Llama 4
Maverick~\cite{llama3_Dubey} via the watsonx.ai API~\cite{ibm_infra}; any
competent instruction-tuned LLM can be used.

\begin{figure}[t]
  \centering
  \includegraphics[width=\linewidth]{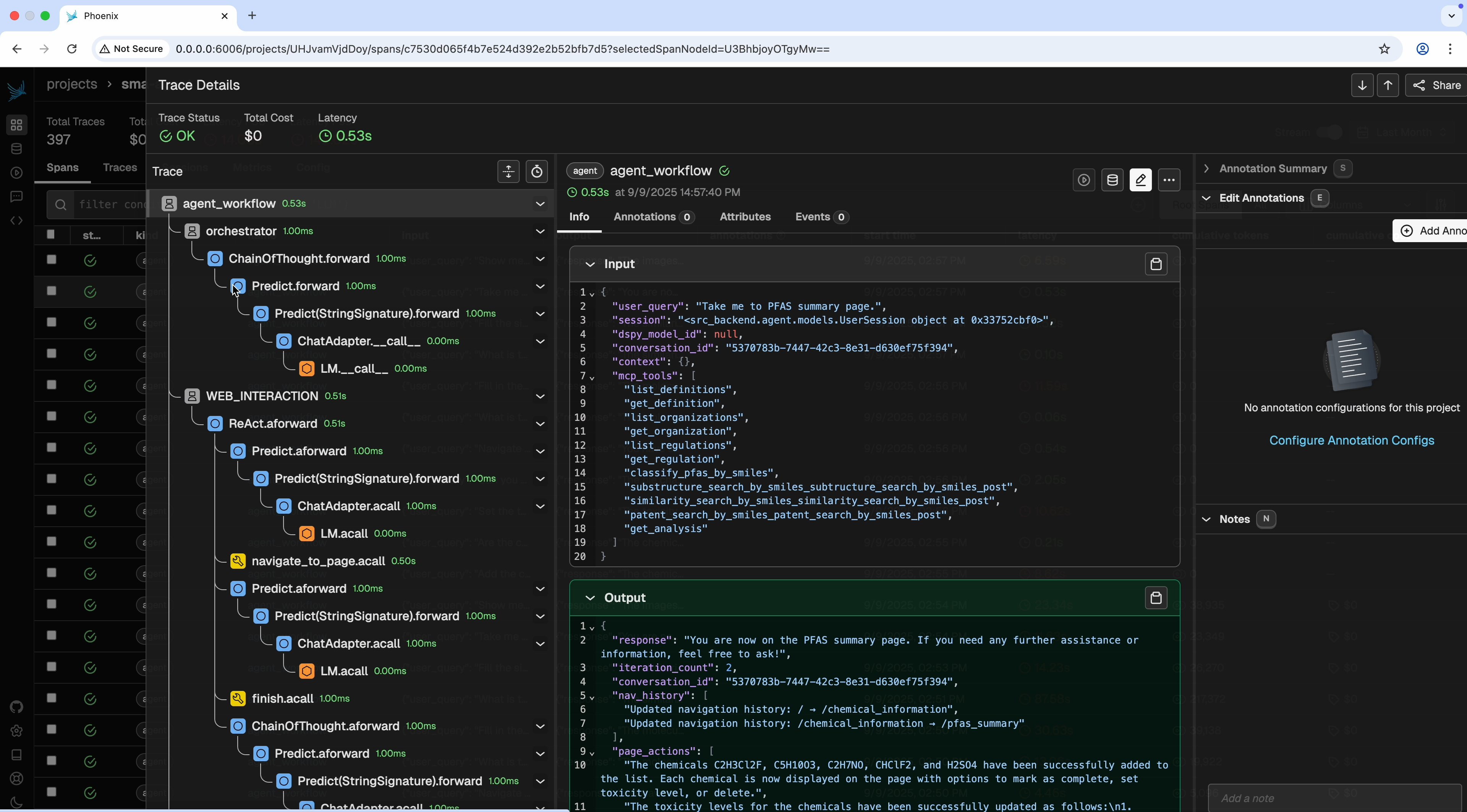}
  \vspace{-4mm}
  \caption{\textbf{Testing interface.} Integration tests evaluate action correctness and latency under simulated frontend-backend interaction.}
  \label{fig:test}
  \vspace{-0.7em}
\end{figure}

\section{Evaluation}

\begin{table}[t]
    \setlength{\abovecaptionskip}{0.1cm}
    \footnotesize
    \caption{\textbf{Integration testing for the chemistry application.}}
    \label{tab:tests}
    \begin{adjustbox}{width=1.0\linewidth,center}
    \begin{tabular}{@{}>{\raggedright\arraybackslash}p{0.38\linewidth}
                      >{\raggedright\arraybackslash}p{0.58\linewidth}@{}}
        \toprule
        \textbf{Target / Metric} & \textbf{Validation / Result} \\
        \noalign{\vskip 0.3ex}
        \hline \noalign{\vskip 1.0ex}
        Web interaction &
        Function names, parameters, and navigation destinations \\
        \noalign{\vskip 0.5ex}
        Analysis and chat &
        Exact response matches or LLM-judged semantic similarity \\
        \noalign{\vskip 0.5ex}
        Multi-turn &
        Action and response sequences across successive queries \\
        \noalign{\vskip 0.5ex}
        Concurrent sessions &
        Session-scoped state and latency inspection via tracing \\
        \noalign{\vskip 0.8ex}
        \hline \noalign{\vskip 1.0ex}
        End-to-end latency$^{\dagger}$ & $14.08$\,s (median) \\
        \noalign{\vskip 0.5ex}
        Execution latency$^{\S}$ & $1.52$\,s (median) \\
        \noalign{\vskip 0.5ex}
        Reported test pass rate$^{\ddagger}$ & $\approx 0.90$ \\
        \bottomrule
    \end{tabular}
    \end{adjustbox}
    \par\smallskip
    \begin{minipage}{\linewidth}
    \scriptsize
    $^{\dagger}$Project-level median latency reported by the archived Phoenix dashboard, which contained 397 traces at the time of capture. These reflect a mixed development workload, including stress tests with unsupported requests, failed cases, and periods of unstable watsonx.ai API connectivity, rather than a standardized task benchmark.
    $^{\S}$Median over sessions replayed with cached LLM responses (i.e., the time spent outside LLM inference).
    $^{\ddagger}$Approximate pass rate reported during development.
    \end{minipage}
    \vspace{-0.5em}
\end{table}

\paragraph{Browser-Free Integration Tests.} The backend accepts textual observations and returns function calls, allowing its integration tests to run without browser automation. We implement the suite using \texttt{pytest}, \texttt{pytest-asyncio}, and \texttt{assertpy}. Each test opens a WebSocket, sends a predefined frontend observation, issues chat queries, and validates the returned actions or responses. Web interaction agents are tested by verifying function calls with correct parameters or navigation to target URLs. Analysis and chat agents are evaluated via exact match or LLM-judged semantic similarity. Multi-turn tests simulate complex end-to-end workflows. Phoenix tracing monitors performance, latency, and session isolation under concurrent sessions. Fig.~\ref{fig:test} illustrates the trace-inspection interface. Table~\ref{tab:tests} summarizes the test targets, trace latencies, and the approximate integration-test pass rate for the chemistry application.

\paragraph{Latency Analysis.} Requests took a median of 14.08\,s end to end (Table~\ref{tab:tests}). The traces were collected against the live application and an internal watsonx.ai endpoint rather than in a controlled benchmark. They include deliberately difficult stress tests, such as unsupported requests, and periods of unstable API connectivity that slowed LLM calls, so the median likely overstates routine latency. Requests in our sessions chain several reasoning steps, tool calls, and interface actions, for example adding five chemicals to a list or classifying each of them as PFAS with reasons. Replaying scripted sessions with cached LLM responses took a median of 1.52\,s per request, suggesting that LLM inference accounts for most of the end-to-end time. Latency should therefore drop substantially with a locally hosted model or a production-grade inference service.

\paragraph{Lessons Learned.} Three details matter more than we expected. (i) Observations must be captured after the DOM settles, because intermediate render states give the agent a stale view. (ii) Agent actions must update both the visible interface and the application's internal state. Browser-automation agents achieve this by simulating real user input, whereas the shim changes values directly. In frameworks such as React, this may leave the underlying application data unchanged, so subsequent operations still use the old value. The shim therefore dispatches the corresponding input events so that the application registers the change. (iii) With our open-weight models, function calling degraded as chat history grew, so the backend keeps only the last $n$ turns; harness engineering can preserve more context but is not the focus of this work.

\section{Limitations and Future Work}

\sys{} depends on accurate ARIA labels and an up-to-date page-function map, both of which are currently maintained manually. Although we integrate \sys{} with two frontend stacks, we have validated it functionally in only one application domain and have not benchmarked it against other agents. Two design properties help mitigate the integration and maintenance burden. First, ARIA labels written according to W3C authoring practices also benefit screen-reader users, meaning that labeling for the agent overlaps with existing accessibility efforts. Second, the function registry describes actions through explicit names, purpose descriptions, and parameter schemas, allowing the reusable backend workflow to consume application-specific actions through a common interface.

Future work includes (i) standardizing ARIA label formats and linting conventions, (ii) semi-automatically extracting candidate actions from the frontend codebase, (iii) developing adapters that allow external agents to discover and invoke registered actions, and (iv) evaluating \sys{} across additional application domains and against existing agents.


\bibliographystyle{named}
\bibliography{ijcai26}

\end{document}